\documentclass[11pt]{article}
\usepackage[preprint]{acl}
\usepackage{times}
\usepackage{latexsym}
\usepackage[T1]{fontenc}
\usepackage[utf8]{inputenc}
\usepackage{microtype}
\usepackage{inconsolata}
\usepackage{graphicx}
\usepackage{multirow}
\usepackage{amsmath}
\usepackage{amssymb}

\title{From Oracle to Noisy Context: Mitigating Contextual Exposure Bias in Speech-LLMs}

\author{
 \textbf{Xiaoyong Guo\textsuperscript{1}},
 \textbf{Nanjie Li\textsuperscript{1}},
 \textbf{Zijie Zeng\textsuperscript{3}},
 \textbf{Kai Wang\textsuperscript{1}},
\\
 \textbf{Hao Huang*\textsuperscript{1,2}},
 \textbf{Haihua Xu\textsuperscript{3}},
 \textbf{Wei Shi\textsuperscript{3}}
\\
\\
 \textsuperscript{1}School of Computer Science and Technology, Xinjiang University, Urumqi, China
\\
 \textsuperscript{2}Joint International Research Laboratory of Silk Road Multilingual Cognitive Computing, Urumqi, China
\\
 \textsuperscript{3}Timekettle
\\
 \small{
   \textbf{Correspondence:} \href{hwanghao@gmail.com}{hwanghao@gmail.com}
 }
}

\begin{document}
\maketitle
\begin{abstract}
Contextual automatic speech recognition (ASR) with Speech-LLMs is typically trained with oracle conversation history, but relies on error-prone history at inference, causing a train–test mismatch in the context channel that we term contextual exposure bias. We propose a unified training framework to improve robustness under realistic histories: (i) Teacher Error Knowledge by using Whisper large-v3 hypotheses as training-time history, (ii) Context Dropout to regularize over-reliance on history, and (iii) Direct Preference Optimization (DPO) on curated failure cases. Experiments on TED-LIUM 3 (in-domain) and zero-shot LibriSpeech (out-of-domain) show consistent gains under predicted-history decoding. With a two-utterance history as context, SFT with Whisper hypotheses reduce WER from 5.59\% (oracle-history training) to 5.47\%, and DPO further improves to 5.17\%. Under irrelevant-context attacks, DPO yields the smallest degradation (5.17\% $\rightarrow$ 5.63\%), indicating improved robustness to misleading context. Our code and models are published on 
\url{https://github.com/XYGuo1996/Contextual_Speech_LLMs}.

\end{abstract}

\section{Introduction}


Automatic Speech Recognition (ASR) has evolved rapidly \cite{ma2025speech}, progressing from traditional architectures like CTC \cite{graves2006connectionist}, AED \cite{chan2016listen}, and RNN-T \cite{graves2013speech} to robust pre-trained models such as HuBert \cite{hsu2021hubert}, Whisper \cite{radford2023robust}, and Whistle \cite{yusuyin2025whistle, li2022recent}. While the recent emergence of Speech-LLMs \cite{cui2024recent} has further expanded multimodal capabilities, effectively incorporating contextual cues---a problem known as Contextual ASR \cite{aleksic2015bringing, hall2015composition}---remains a critical challenge across these paradigms to compensate for ambiguous acoustic evidence.

In conventional ASR systems, the utilization of contextual information primarily follows two core paradigms: Shallow Fusion and Deep Fusion \cite{fang2025joint}. Shallow Fusion functions essentially as an inference-stage ensemble strategy. Enhances the accuracy of recognition within specific contexts by utilizing an external language model to score the hypotheses generated by the acoustic model \cite{mcdermott2019density, ravi2020improving, guo2023improved}. In contrast, Deep Fusion involves a more fundamental architectural innovation. This approach encodes contextual information into vector embeddings that are directly integrated into the internal components of end-to-end models. Consequently, this enables the model to learn joint representations of acoustic features and contextual cues during the training phase \cite{toshniwal2018comparison, wang2024deep, tang2024improving, huang2024optimizing, huang2024improving, kolokolov2024self, shi2024seaco, sudo2024contextualized}. Such as Contextual RNN-T \cite{jain2020contextual} demonstrated the effectiveness of attending to metadata embeddings for rare word recognition, while Hou et al. \cite{hou2022bring} extended this by integrating dialogue history into streaming RNN-T encoders. Hori et al. \cite{hori2020transformer} explored Transformer-based architectures for long-context modeling, highlighting the importance of cross-utterance dependencies. 

Benefiting from the robust contextual reasoning of LLMs, recent research has prioritized embedding contextual cues into Speech-LLMs prompts \cite{chen2024salm, yang2024mala, shen2025retrieval, lakomkin2024end, cheng2024context, lei2025contextualization, koshkin2024llms, fang2025joint, gong2024contextual, zhou2025boosting}. Approaches vary from using textual metadata like titles and descriptions \cite{lakomkin2024end} to incorporating specific entity lists \cite{chen2024salm} and multi-modal auxiliary inputs \cite{yang2024mala}. Notably, Lakomkin et al \cite{lakomkin2024end} also examined the model's resilience to contextual perturbations. Building on this trend, Retrieval-Augmented Generation (RAG) has been proposed as a novel method to integrate Speech-LLMs \cite{shen2025retrieval, li2024rag, mu2025hearing, gourav2024multi}. On the other hand, Step-Audio employs a text-based context manager to maintain conversation history and support multi-turn interactions \cite{huang2025step}. The paper does not provide a separate analysis of the impact of historical text on system performance.


We study utterance-by-utterance contextual ASR, where each utterance is decoded sequentially using transcripts from previous turns as textual history. In the training stage, the model can rely on oracle history, but in deployment, oracle history is unavailable and the model must condition on error-prone ASR hypotheses, yielding a distribution shift between training-time and inference-time context. We call this mismatch in the context channel contextual exposure bias.
While the challenge of utilizing imperfect context has been approached—for instance, via context dropout in speech translation \cite{hussein2024enhancing} or noise representation learning in dialogue ASR \cite{lee2024enhancing} these methods often rely on implicit regularization or auxiliary modules. They lack a direct mechanism to align the model’s generation preferences to explicitly reject contextual errors.

To mitigate exposure bias in continuous-utterance ASR, we propose a unified training framework that integrates three complementary strategies.  Teacher Error Knowledge: instead of using ground-truth history, we feed Whisper large-v3 decoding hypotheses as training context, exposing the model to realistic “teacher errors” and better matching inference-time conditions. Context Dropout: we randomly mask the historical context with a fixed probability, reducing over-reliance on text history and encouraging acoustic-focused transcription, thereby alleviating history overfitting. Direct Preference Optimization (DPO): we further refine contextual generation by constructing preference pairs from selected hard negatives, explicitly training the model to avoid negative behaviors and improve output quality. The main contributions of this paper are summarized as follows:
\begin{itemize}
  \item \textbf{Leveraging contextual exposure bias to improve Speech-LLMs ASR:} 
  We identify contextual exposure bias as a key failure mode in context-conditioned Speech-LLMs based ASR and,  explicitly exploit this train-test mismatch as a guiding principle to design training and alignment strategies that boost recognition performance under realistic, imperfect historical context.


  \item \textbf{Noise-aware training for imperfect context:} 
  We propose a unified training framework that aligns training with inference by (i) using a strong teacher ASR system to generate realistic, error-prone historical context during training (instantiated with Whisper large-v3 in our experiments), (ii) applying context dropout to regularize reliance on history, and (iii) using DPO with preference pairs constructed from contextual failure cases to reduce error amplification.

  \item \textbf{Robustness under realistic and cross-domain evaluation:} 
  Training with strong teacher-generated historical context substantially closes the oracle--inference gap, and DPO further improves robustness by reducing the model’s sensitivity to irrelevant or erroneous context, consistently yielding the best in-domain and cross-domain performance under realistic decoding conditions.

\end{itemize}

\section{Method}
\subsection{Utterance-level Contextual ASR}
\label{section: Utterance-level Contextual ASR}
We consider an utterance-level contextual ASR setting, in which each utterance is recognized sequentially and the transcripts of preceding utterances are provided as textual context for the current decoding.

Formally, an input stream is segmented into a sequence of utterances $\{X_{t}\}_{t=1}^{T}$, where $X_t$ denotes the acoustic signal of the $t$-th utterance, and $Y_t$ denotes its reference transcript. For each utterance $t$, the recognizer produces a hypothesis $\hat{Y}_t$ conditioned on the current acoustics $X_t$ and an available context $C_t$:
\begin{equation}
    \hat{Y}_t \sim p_{\theta}(Y|X_t,C_t)
\end{equation}
We define the textual context $C_t$ as a function of preceding utterances' transcripts. In the simplest case, $C_t$ is the concatenation of the most recent $N$ utterance-level transcripts:
\begin{equation}
    C_t = concat(S_{t-N}, \cdots, S_{t-1})
\end{equation}
where $S_i$ is the transcript used as history for utterance $i$, and $N$ controls the context window size (with $N=0$ corresponding to the no-context baseline).

A central aspect of this setting is that the quality of historical transcripts available at inference may differ from that used during training. In many experimental setups, $S_i$ is taken to be the oracle transcript $Y_i$, yielding an oracle-context condition. In realistic deployment, however, the system does not have access to $Y_i$ and must instead rely on automatically generated hypotheses from upstream recognition of previous turns. We denote this predicted-context condition by setting $S_i=\hat{Y}_i$, where $\hat{Y}_i$ may contain recognition errors, relative to reference ${Y_i}$. 

This discrepancy induces a distribution mismatch in the conditioning context between training and inference: training commonly conditions on oracle history, while inference conditions on imperfect, model-generated history. In this work, we refer to this train–test mismatch in the contextual channel as contextual exposure bias. Importantly, this is a context-level mismatch: even if the model is trained with standard teacher-forcing for the current utterance, the historical context provided to the model at inference can be error-prone and may systematically differ from the oracle context used in training.

In the remainder of the paper, we study how this contextual exposure bias affects contextual ASR in Speech-LLMs, and we develop training strategies that better align training-time context with inference-time conditions.

\subsection{Model Architecture}

\begin{figure}
    \centering
    \includegraphics[width=0.8\linewidth]{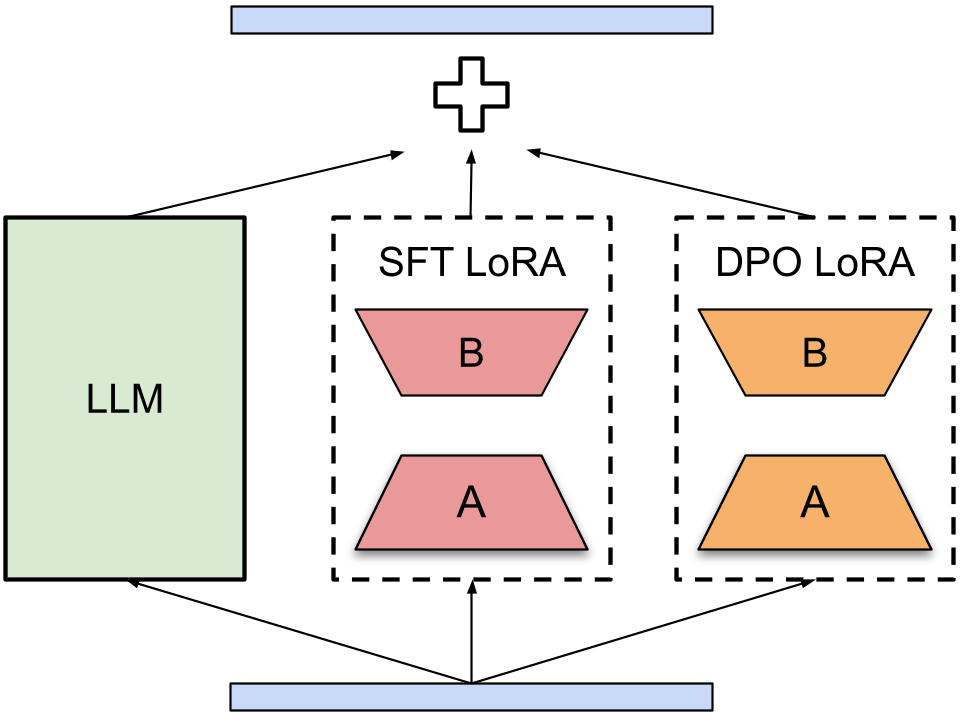}
    \caption{Model architecture}
    \label{fig:model-arch}
\end{figure}

We adopt a training framework consisting of an off-the-shelf speech encoder, a frozen LLM backbone, and a trainable MLP projector for modality adaptation. We fine-tune the LLM with Low-Rank Adaptation (LoRA) \cite{hulora} in two stages: (i) a dedicated SFT LoRA module for SFT, and (ii) a separate DPO LoRA module for preference alignment. Following \cite{jiang2025huvidpo}, the DPO stage introduces an additional, distinct LoRA block rather than reusing the SFT adapter, enabling independent control of preference optimization. The overall architecture is illustrated in Figure~\ref{fig:model-arch}.

\subsection{Training Methodology}
We study contextual ASR under a realistic train--test discrepancy: during training, the contextual history is typically provided as oracle transcripts, whereas at inference time the history must be obtained from upstream ASR outputs and therefore contains recognition errors. To bridge this gap, we propose a training framework that (i) replaces oracle history with teacher-generated hypotheses to simulate imperfect context during training, where the teacher is a strong off-the-shelf ASR system (instantiated as Whisper large-v3 in our experiments), and (ii) regularizes the model's reliance on history to prevent error amplification when the context is noisy or misleading.


\subsubsection{Teacher Error Knowledge}
To approximate inference-time conditions, we introduce Teacher Error Knowledge, where the history transcripts $\hat{Y}$ are taken from Whisper large-v3 decoded hypotheses instead of always using oracle history \footnote{Our method does not rely on a specific teacher model.}. Concretely, we pre-decode the training set with Whisper and store hypotheses $\hat{Y}_i^{Whisper}$. During training, we construct $C_t$ using Whisper large-v3 decoded hypotheses as the history source. This exposes the model to realistic contextual noise and encourages robustness to history errors.

We optimize the standard sequence negative log-likelihood conditioned on the constructed context:
\begin{equation}
    \mathcal{L}_{SFT} = - \sum_{t} \log{p_{\theta}(Y_t|X_t, C_t, P)}
\end{equation}
where $P$ is the task prompt and $\theta$ denotes trainable parameters.

\subsubsection{Context Dropout}
Even with noisy histories, the model may still over-trust contextual text and become brittle when the history is misleading.To further regularize context usage, we apply Context Dropout during training: with probability $p_{drop}$, we mask the textual history $C_t$, while keeping the current utterance speech $X_t$ unchanged:
\begin{equation}
    \tilde{C}_{t} = \begin{cases}
        \varnothing, & \text{with~} p_{\text{drop}}, \\
        C_{t}, & \text{otherwise}
    \end{cases}
\end{equation}
The SFT objective is then computed conditioned on $\tilde{C}_t$. This mechanism prevents the model from depending on history as a shortcut and encourages it to retain strong speech information, while still allowing it to benefit from context when available.

\subsubsection{Preference Optimization on Hard Negatives with DPO}
While SFT with noisy history improves overall robustness, we observe that a small subset of failure cases can still be triggered. To explicitly suppress such undesirable behaviors, we further apply DPO on curated hard negatives.

For each selected utterance, we construct a preference pair $(Y^+, Y^-)$ under the same inputs $\mathbf{X}=(X_t, C_t, P)$, where $Y^+$ is the ground-truth and $Y^-$ is the inference transcript of the model. DPO optimizes the policy to assign higher likelihood to $Y^+$ than $Y^-$ without requiring an explicit reward model. Following the standard DPO formulation, we minimize:
\begin{equation}
\Delta_{\theta}
= \log \pi_{\theta}(Y^{+}\mid \mathbf{X}) - \log \pi_{\theta}(Y^{-}\mid \mathbf{X})
\end{equation}
\begin{equation}
\Delta_{r}= \log \pi_{r}(Y^{+}\mid \mathbf{X})-\log \pi_{r}(Y^{-}\mid \mathbf{X})
\end{equation}
\begin{equation}
m = \beta \left(\Delta_{\theta} - \Delta_{\mathrm{r}}\right)
\end{equation}
\begin{equation}
\mathcal{L}_{\mathrm{DPO}} = -\log \sigma(m)
\end{equation}
In our experiments, we set the reference policy $\pi_r$ to the SFT checkpoint (frozen), $\beta$ is a temperature coefficient that controls the strength of preference sharpening, in this work, we use $\beta=0.1$, and $\sigma(\cdot)$ is the sigmoid function. Here, $\Delta_{\theta}$ and $\Delta_{r}$ measure the log-likelihood gaps between the preferred and dispreferred responses under the current policy $\pi_{\theta}$ and the reference policy $\pi_r$, respectively, and $m$ scales their difference to form the DPO objective \cite{ouyang2022training, rafailov2023direct}.

In our implementation, we use a separate LoRA block for DPO to decouple preference optimization from the main SFT adaptation \cite{jiang2025huvidpo} as in Figure \ref{fig:model-arch}.

\section{Experiment Setup}
\subsection{Models and Modules}
We build our system on an in-house Speech-LLMs training framework. We adopt the Whisper large-v3 \cite{radford2023robust}, discard the decoder and only use the encoder as a feature extractor. Based on previous work on Speech-LLMs, which has shown that LLMs with SFT significantly outperform vanilla pre-trained models in speech recognition tasks \cite{ma2025speech}, we choose vicuna-7B-v1.5 \cite{zheng2023judging} with LoRA adapter as the LLM for our system. Unless otherwise specified, we keep the Whisper large-v3 encoder frozen and train a lightweight MLP projector together with LoRA parameters. We report WER as the primary metric.

\subsection{Datasets}
We conduct both training and in-domain evaluation on the TED-LIUM 3 \cite{hernandez2018ted} dataset following an utterance-level segmentation. Each utterance is paired with its reference transcript, and contextual history is constructed from preceding utterances within the same session. We use the official splits for train, dev and test.

To assess cross-domain generalization, we evaluate on LibriSpeech \cite{panayotov2015librispeech} without any additional training. We report results on test-clean and test-other.

\subsection{Training Protocol}
We train our system in three stages. All experiments are conducted on 4 NVIDIA A100 GPUs. In the first Stage, we train a context-free base model by freezing the Whisper large-v3 encoder and Vicuna v1.5-7B, and optimizing only the MLP projector with a standard ASR objective. This stage establishes a reliable alignment between speech and text, which serves as a stable initialization for subsequent contextual training. For this stage, the batch size is set to 6, the learning rate to $1\text{e-}4$, and the warmup step to 1000.


In the second stage, we perform SFT of the Speech-LLMs using paired speech–text data under the \nameref{section: Utterance-level Contextual ASR} setting.  During SFT, we replace oracle history with these Whisper hypotheses to expose the model to history noise. Importantly, Whisper is only used offline to construct Teacher Error Knowledge for training; our inference doesn't require Whisper as an auxiliary component. To further reduce over-reliance on the history channel, we apply Context Dropout, which randomly masks the provided textual history with probability $p=0.5$ (and keeps the rest of the training pipeline unchanged). This regularizer forces the model to remain grounded in acoustic information when the history is missing or unreliable. In this stage, the batch size is set to 6, the learning rate is $1\text{e-}4$, and the warmup step is 1000.

In the third stage, we refine the model on a subset of challenging examples, termed 'hard negatives', which are identified by decoding the training set using the optimal model from Stage 2 and selecting instances with a WER exceeding 20\%. On this specific subset, we conduct a comparative analysis of two refinement strategies: an additional SFT pass and DPO, where the latter utilizes preference pairs to explicitly mitigate error propagation under noisy-history conditions. To implement these strategies, we introduce a dedicated LoRA module distinct from the previous SFT phase while keeping the model backbone frozen. In this stage, for the additional SFT, the batch size is 6; for DPO, the batch size is 2, the accumulate grad is 16. Both strategies use 0 warmup steps and learning rate is $1\text{e-}5$.  

\subsection{Inference Details}
During the inference phase, we employ beam search decoding to balance generation quality and diversity. Specifically, we set the beam width to 4 and the maximum generation length to 200 tokens. To ensure deterministic and reproducible evaluation, we disable sampling strategies (setting do\_sample=False, top\_p=1.0, and temperature=1.0). We also maintain neutral penalty settings with both repetition\_penalty and length\_penalty set to 1.0.

\paragraph{Inference-time LoRA strength adjustment.}
We introduce an explicit inference-time strength factor $\gamma$ to control the contribution of the DPO LoRA adapter, while keeping the LoRA rank and scaling hyperparameters fixed across training and inference.

\paragraph{LoRA formulation.}
Let $W_{\text{LLM}}$ denote the frozen base LLM weights. We attach two LoRA adapters: an SFT adapter and a DPO adapter. The resulting effective weight matrix used in the forward pass is:
\begin{equation}
    W=W_{\text{LLM}}
    +\frac{\alpha}{r}\,\Delta W_{\text{SFT}}
    +\gamma\,\frac{\alpha^\prime}{r^\prime}\,\Delta W_{\text{DPO}}
\end{equation}
where $\Delta W_{\text{SFT}}$ and $\Delta W_{\text{DPO}}$ are low-rank updates (e.g., $\Delta W = BA$) learned in the SFT and DPO stages, respectively, and $\frac{\alpha}{r}$ (resp., $\frac{\alpha^\prime}{r^\prime}$) controls the magnitude of the injected LoRA features \cite{hulora}.
In our implementation, we use the same rank and scaling for both adapters, i.e.,
\begin{equation}
    r = r^\prime,\qquad \alpha = \alpha^\prime
\end{equation}
specifically, $r=8$ and $\alpha=32$, so the only additional degree of freedom for the DPO adapter at inference is $\gamma$.

\paragraph{Training vs. inference.}
During DPO training, we set $\gamma=1$, so the DPO adapter is applied with its full strength:
\begin{equation}
    W_{\text{train}}=W_{\text{LLM}}+\frac{\alpha}{r}\,\Delta W_{\text{SFT}}
    +\frac{\alpha}{r}\,\Delta W_{\text{DPO}}
\end{equation}
At inference, we keep $r$ and $\alpha$ unchanged and instead tune $\gamma<1$ to soften the effect of the DPO adapter:
\begin{equation}
    W_{\text{infer}}=W_{\text{LLM}}+\frac{\alpha}{r}\,\Delta W_{\text{SFT}}
    +\gamma\,\frac{\alpha}{r}\,\Delta W_{\text{DPO}}
\end{equation}
In our experiments, we set $\gamma=0.25$, which effectively reduces the DPO adapter strength by $4\times$ compared to training, while leaving the learned parameters and all other LoRA hyperparameters intact.

\paragraph{Why adjust $\gamma$ at inference.}
Empirically, we observe that the DPO-tuned adapter can be overly aggressive during decoding, occasionally exhibiting reward over-optimization (i.e., the model exploits preference signals at the expense of coherent and faithful generation) \cite{gao2023scaling}. Introducing $\gamma$ provides a simple and stable knob to control the influence of DPO at test time: lowering $\gamma$ mitigates over-optimization and helps preserve the generalization and fluency of the underlying base model, while retaining most of the robustness gains from preference alignment.

\section{Result and Analysis}
\subsection{Main Result}

We verify the effectiveness of our proposed framework on the TED-LIUM 3 dataset. Table \ref{tab:Main Result} presents the WER comparisons between baselines and our proposed methods. To ensure a fair comparison under realistic deployment conditions, we primarily focus on the settings where $\text{Con}_{\text{inf}}$ is hypotheses, meaning the model must rely on predicted history rather than oracle transcripts.

\begin{table*}[tp]
    \centering
    \setlength{\tabcolsep}{3pt}
    \begin{tabular}{ll|cccc|cccc}
        \hline
        \multirow{2}{*}{\textbf{$N$}} & \multirow{2}{*}{\textbf{$\text{Con}_{\text{inf}}$/$\text{Con}_{\text{train}}$}} & \multicolumn{4}{c}{\textbf{0 Dropout WER (\%)$\downarrow$}} & \multicolumn{4}{|c}{\textbf{0.5 Dropout WER (\%)$\downarrow$}} \\
        & & \textbf{TED} & \textbf{Test-clean} & \textbf{Test-other} & \textbf{LS-Ave.} & \textbf{TED} & \textbf{Test-clean} & \textbf{Test-other} & \textbf{LS-Ave.} \\
        \hline
        0 & - / - & 7.89 & 4.79 & 9.83 & 7.310 & - & - & - & - \\
        \hline
        \multirow{5}{*}{1} & GT / GT & 5.6 & 4.49 & 10.36 & 7.425 & 7.89 & 4.31 & 9.68 & 6.995 \\
        & hyp / GT & 5.85 & \textbf{4.54} & 10.63 & 7.585 & 7.47 & \underline{4.74} & 9.94 & 7.340 \\
        & hyp / Whisper & \textbf{5.62} & \underline{4.67} & \textbf{9.46} & \textbf{7.065} & \underline{7.21} & 5.37 & 9.96 & 7.665 \\
        & \hspace{1em} + DPO & \underline{5.69} & 4.71 & 9.57 & 7.140 & \textbf{5.32} & \textbf{4.56} & \underline{9.38} & \textbf{6.970} \\
        & \hspace{1em} + SFT2 & 5.76 & \underline{4.67} & \underline{9.49} & \underline{7.080} & 7.26 & 5.14 & \textbf{9.30} & \underline{7.220} \\
        \hline
        \multirow{5}{*}{2} & GT  / GT & 6.73 & 4.10 & 8.36 & 6.230 & 5.66 & 4.10 & 8.37 & 6.235 \\
        & hyp / GT & \underline{6.89} & \underline{4.85} & \underline{9.88} & \underline{7.365} & 5.59 & 5.15 & \textbf{9.10} & \underline{7.130} \\
        & hyp / Whisper & 8.15 & 5.57 & 12.00 & 8.785 & \underline{5.47} & \underline{5.14} & 9.50 & 7.320 \\
        & \hspace{1em} + DPO & \textbf{5.07} & 4.87 & \textbf{9.51} & \textbf{7.190} & \textbf{5.17} & \textbf{4.84} & \underline{9.19} & \textbf{7.015} \\
        & \hspace{1em} + SFT2 & 6.90 & \textbf{4.55} & 11.17 & 7.860 & 6.10 & 5.43 & 9.66 & 7.545 \\
        \hline
        \multirow{5}{*}{3} & GT  / GT & 7.35 & 4.24 & 8.29 & 6.265 & 10.42 & 4.89 & 10.36 & 7.625 \\
        & hyp / GT & \underline{7.05} & \textbf{5.03} & 10.68 & \underline{7.855} & 12.62 & \underline{5.28} & 10.93 & \underline{8.105} \\
        & hyp / Whisper & 10.06 & 5.36 & 10.69 & 8.025 & \underline{7.87} & 5.93 & 10.39 & 8.160 \\
        & \hspace{1em} + DPO & \textbf{5.98} & 5.60 & \textbf{9.96} & \textbf{7.780} & \textbf{5.18} & \textbf{4.73} & \textbf{9.36} & \textbf{7.045} \\
        & \hspace{1em} + SFT2 & 9.30 & \underline{5.22} & \underline{10.49} & \underline{7.855} & 8.01 & 6.11 & \underline{10.20} & 8.155 \\
        \hline
        \multirow{5}{*}{4} & GT  / GT & 8.54 & 4.26 & 9.01 & 6.635 & 9.22 & \textbf{4.75} & 10.23 & 7.490 \\
        & hyp / GT & \underline{7.74} & 4.87 & 11.07 & 7.970 & 10.87 & \textbf{4.75} & 10.23 & 7.490 \\
        & hyp / Whisper & 87.37 & \textbf{4.66} & \underline{10.81} & \underline{7.735} & \underline{7.81} & \textbf{4.75} & \underline{9.82} & \underline{7.285} \\
        & \hspace{1em} + DPO & \textbf{4.93} & \underline{4.79} & \textbf{9.97} & \textbf{7.380} & \textbf{5.69} & \underline{4.79} & \textbf{9.25} & \textbf{7.020} \\
        & \hspace{1em} + SFT2 & 113.95 & 4.90 & 11.34 & 8.120 & 9.16 & \textbf{4.75} & 9.83 & 7.290 \\
        \hline
        \multirow{5}{*}{5} & GT  / GT & 8.72 & 5.46 & 9.49 & 7.475 & 9.57 & 4.90 & 10.04 & 7.470 \\
        & hyp / GT & \underline{10.34} & 5.08 & 10.76 & 7.920 & \underline{8.19} & 5.36 & 11.29 & 8.325 \\
        & hyp / Whisper & 135.57 & \textbf{4.59} & 10.87 & \underline{7.730} & 8.5 & \underline{4.95} & \underline{9.33} & \underline{7.140} \\
        & \hspace{1em} + DPO & \textbf{5.34} & \underline{4.67} & \textbf{9.85} & \textbf{7.260} & \textbf{4.96} & \textbf{4.55} & \textbf{9.24} & \textbf{6.895} \\
        & \hspace{1em} + SFT2 & 72.55 & 4.90 & \underline{10.57} & 7.735 & 8.51 & 5.23 & \underline{9.33} & 7.280 \\
        \hline
    \end{tabular}
    \caption{WER comparison on TED-LIUM 3 and out-of-domain Librispeech dataset across different context window sizes (\textbf{$N$}). The column \textbf{$\text{Con}_{\text{inf}}$ / $\text{Con}_{\text{train}}$} specifies the source of history used during inference and training, respectively. \textbf{hyp} denotes using the model's own predictions as history during inference. Regarding training configuration, \textbf{GT} uses ground-truth history, while \textbf{Whisper} indicates the model was trained using context decoded by Whisper to simulate historical errors. \textbf{+ DPO} and \textbf{+ SFT2} are additional fine-tuning stages applied to the SFT model.}
    \label{tab:Main Result}
\vspace{0.5cm}
    \centering
    \begin{tabular}{l|ccc|ccc}
        \hline
         \multirow{2}{*}{\textbf{$\gamma$}} & \multicolumn{3}{c|}{\textbf{TED-LIUM 3 (WER \%) $\downarrow$}} & \multicolumn{3}{c}{\textbf{LibriSpeech (WER \%) $\downarrow$}} \\
         & \textbf{Attacks/o} & \textbf{Attacks/w} & \textbf{Gap $\downarrow$} & \textbf{Test-clean} & \textbf{Test-other} & \textbf{Ave.} \\
        \hline
        0 & 5.47 & 7.93 & 2.46 & 5.14 & 9.50 & 7.320 \\
        0.0625  & 5.37 & 7.13 & 1.76 & 5.12 & 9.31 & 7.215 \\
        0.125  & 5.11 & 5.76 & 0.65 & 5.02 & 9.53 & 7.275 \\
        0.1875  & \textbf{5.06} & 5.69 & 0.63 & \textbf{4.70} & \textbf{9.08} & \textbf{6.890} \\
        0.25  & 5.17 & \textbf{5.63} & 0.46 & 4.84 & 9.19 & 7.015 \\
        0.375 & 5.55 & 5.73 & \textbf{0.18} & 4.85 & 9.63 & 7.240 \\
        0.5 & 8.39 & 8.67 & 0.28 & 6.44 & 12.14 & 9.290 \\
        0.625 & 53.26 & 57.15 & 3.89 & 27.11 & 28.96 & 28.035 \\
        \hline
    \end{tabular}
    \caption{Impact of DPO LoRA scaling factor ($\gamma$) during inference. TED-LIUM 3 Gap denotes the WER degradation caused by irrelevant context attacks. ``Attacks/o" refers to relevant context inference, while ``Attacks/w" refers to irrelevent context randomly selected from the test set.  }
    \label{tab:DPO alpha}
\end{table*}

\textbf{Impact of Context and Training Strategies}: The comparison with the baseline without context (7.89\%) reveals that incorporating context does not inherently guarantee performance improvements; it depends heavily on the training strategy. As shown in Table \ref{tab:Main Result}, the configuration with $\text{Con}_{\text{train}}$ as Whisper large-v3 generated context without context dropout (0 Dropout) yields a WER of 8.15\% at $N=2$, which underperforms the context-free baseline. This degradation indicates that without regularization, the model may overfit to noisy history or become overly reliant on it. In contrast, the model trained with 0.5 Dropout effectively leverages the context. Under the same configuration ($N=2$, Whisper history), it achieves a WER of 5.47\%, significantly surpassing the No Context baseline. Consequently, we identify the model trained with Whisper large-v3 generated context and 0.5 dropout model as our best-performing SFT model, which serves as the robust baseline for subsequent optimization.

\textbf{Sensitivity to Regularization and Exposure Bias:} We analyze how different training data sources affect the model's sensitivity to context dropout. As shown in Table \ref{tab:Main Result}, the baseline trained with ground-truth context ($\text{Con}_{\text{train}} = \text{GT}$) exhibits an inconsistent preference for dropout rates depending on the context length. Specifically, for $N=1, 3, 4$, the model performs better without dropout (0 Dropout), whereas for $N=2, 5$, it requires 0.5 Dropout to achieve optimal results. This fluctuation suggests that models with $\text{Con}_{\text{train}} = \text{GT}$ are structurally unstable, requiring context-length-specific hyperparameter tuning to avoid performance degradation (e.g., using the wrong 0.5 Dropout at $N=3$ leads to a high WER of 12.62\%). In contrast, our model using Whisper training context ($\text{Con}_{\text{train}} = \text{Whisper}$) shows a clear and consistent pattern for sequential tasks. While 0 Dropout is preferred for the shortest context ($N=1$), for all multi-turn scenarios ($N \ge 2$), 0.5 Dropout consistently yields superior performance and is essential for stability. Comparing the optimal configurations, our method ($\text{Con}_{\text{train}} = \text{Whisper}$ with 0.5 Dropout) achieves the best overall performance at $N=2$ (5.47\% vs. GT's best 5.59\%). Furthermore, unlike the ground-truth baseline which oscillates between needing and rejecting regularization, our method provides a reliable strategy ($p=0.5$) that ensures robustness across varying context lengths, avoiding the catastrophic failures seen in unregularized Whisper models.


\textbf{Improvement with DPO:} {\color{red}}To further suppress recognition errors, we refine each SFT checkpoint using the “hard negatives” mined by the same strategy, comparing additional SFT (SFT2) against DPO. Table 1 shows that SFT2 is not a reliable refinement strategy: it often provides negligible gains and frequently degrades performance (e.g., at $N=2$ with 0.5 dropout, TED WER increases from 5.47\% to 6.10\%). In contrast, DPO consistently improves WER across almost all configurations (9 out of 10), with particularly large gains under longer context windows where error accumulation is more severe (e.g., 7.87\%→5.18\% at $N=3$, 0.5 dropout; 8.50\%→4.96\% at $N=5$, 0.5 dropout). We attribute this to the fact that SFT only forces the model to mimic the ground truth, whereas DPO explicitly optimizes the model to prefer the ground truth over its erroneous hypotheses, yielding more stable improvements.\color{black}


\subsection{Cross-Domain Generalization}
{\color{red}}To assess out-of-domain generalization, we evaluate all models on LibriSpeech (test-clean/test-other) in a zero-shot setting using utterance-level decoding, where the inference history is formed by the model’s own hypotheses. As shown in Table \ref{tab:Main Result}, SFT with noisy history does not reliably improve cross-domain transfer: the \textit{hyp/Whisper} SFT model at N=2 with 0.5 dropout achieves 7.32\% LS-Ave., which is essentially on par with the no-context baseline (7.31\%), and increasing the context window can even degrade performance (e.g., 8.16\% at N=3, 0.5 dropout).

In contrast, DPO consistently improves robustness across multiple context window sizes. For example, DPO reduces LS-Ave. from 7.32\% to 7.015\% at N=2 (0.5 dropout), and from 8.16\% to 7.045\% at N=3 (0.5 dropout). The best out-of-domain result is obtained with N=5 and 0.5 dropout, where DPO achieves 6.895\% LS-Ave., outperforming both the no-context baseline and the corresponding SFT baselines. This finding suggests that DPO effectively mitigates domain-specific overfitting. By penalizing the repetition of history errors and hallucinations through preference learning, the model learns to better balance contextual cues with acoustic evidence, leading to superior generalization on unseen out-of-domain data.\color{black}


\begin{table*}[tp]
    \centering
    \begin{tabular}{ll|ccc|ccc}
        \hline
        \multirow{2}{*}{\textbf{$N$}} & \multirow{2}{*}{\textbf{\textbf{$\text{Con}_{\text{inf}}$ / $\text{Con}_{\text{train}}$}}} & \multicolumn{3}{c}{\textbf{0 Dropout WER (\%)$\downarrow$}} & \multicolumn{3}{|c}{\textbf{0.5 Dropout WER (\%)$\downarrow$}} \\
        & & \textbf{Attacks/o} & \textbf{Attacks/w} & \textbf{Gap $\downarrow$} & \textbf{ Attacks/o} & \textbf{Attacks/w} & \textbf{Gap $\downarrow$} \\
        \hline
        \multirow{3}{*}{1} & hyp / GT & 5.85 & 8.32 & 2.47 & 7.47 & 8.82 & 1.35 \\
        & hyp / Whisper & \textbf{5.62} & 6.27 & 0.65 & 7.21 & 7.09 & \textbf{-0.12} \\
        & \hspace{1em} + DPO & 5.69 & \textbf{5.5} & \textbf{-0.19} & \textbf{5.32} & \textbf{5.23} & -0.09 \\
        \hline
        \multirow{3}{*}{2} & hyp / GT & 6.89 & 8.43 & 1.54 & 5.59 & 9.23 & 3.46 \\
        & hyp / Whisper & 8.15 & 10.37 & 2.22 & 5.47 & 7.93 & 2.46 \\
        & \hspace{1em} + DPO & \textbf{5.07} & \textbf{6.59} & \textbf{1.52} & \textbf{5.17} & \textbf{5.63} & \textbf{0.46} \\
        \hline
        \multirow{3}{*}{3} & hyp / GT & 7.05 & 7.64 & 0.59 & 12.62 & 10.19 & \textbf{-2.43} \\
        & hyp / Whisper & 10.06 & 11.18 & 1.12 & 7.87 & 9.53 & 1.66 \\
        & \hspace{1em} + DPO & \textbf{5.98} & \textbf{6.24} & \textbf{0.26} & \textbf{5.18} & \textbf{5.31} & 0.13 \\
        \hline
        \multirow{3}{*}{4} & hyp / GT & 7.74 & 9.75 & 2.01 & 10.87 & 13.15 & 2.28 \\
        & hyp / Whisper & 87.37 & 8.8 & \textbf{-78.57} & 7.81 & 8.82 & \textbf{1.01} \\
        & \hspace{1em} + DPO & \textbf{4.93} & \textbf{5.44} & 0.51 & \textbf{5.69} & \textbf{7.58} & 1.89 \\
        \hline
        \multirow{3}{*}{5} & hyp / GT & 10.34 & 11.83 & 1.49 & 8.19 & 10.41 & 2.22 \\
        & hyp / Whisper & 135.57 & 10.74 & \textbf{-124.83} & 8.5 & 11.34 & 2.84 \\
        & \hspace{1em} + DPO & \textbf{5.34} & \textbf{6.20} & 0.86 & \textbf{4.96} & \textbf{5.51} & \textbf{0.55} \\
        \hline
    \end{tabular}
    \caption{Robustness analysis against irrelevant context attacks on TED-LIUM 3. We evaluate model resilience by replacing the historical context with randomly sampled irrelevant context.}
    \label{tab:TED Attacks}
\end{table*}

\subsection{Impact of DPO Inference Scaling}
\label{sec: Impact of DPO Inference Scaling}
To determine the optimal inference strategy for the DPO module, we analyze the sensitivity of the LoRA scaling factor $\gamma$. As shown in Table \ref{tab:DPO alpha}, there is a critical trade-off derived from the DPO signal strength. While $\gamma=0.1875$ yields the best clean accuracy, increasing $\gamma$ to 0.25 significantly enhances robustness, reducing the performance drop under attack (TED Gap) to 0.46\%. This confirms that a stronger DPO weight helps the model reject misleading context. However, excessive scaling ($\gamma > 0.375$) causes reward over-optimization. We therefore select $\gamma=0.25$ to prioritize stability and error suppression.

\subsection{Robustness to Irrelevant Context}

To verify whether the model genuinely comprehends the contextual information, we conduct a robustness attack test. In this experiment, we replace the ground-truth history with semantically irrelevant context randomly sampled from the TED dataset during inference. A robust contextual ASR model should be able to identify the irrelevance of the context and back off to the acoustic signal, thereby minimizing performance degradation.

{\color{red}}We compare the performance degradation of the proposed method against the standard oracle-trained baseline. As reported in Table \ref{tab:TED Attacks}, the DPO optimized model exhibits remarkable resilience. Across all context window sizes ($N=[1,5]$) and both dropout settings (0 and 0.5), although the degradation gap (Attacks/w $ - $ Attacks/o) of DPO is not always the smallest, the DPO refined model consistently achieves the lowest attacked WER. Overall, these results suggest that incorporating noisy teacher history and context dropout, together with DPO refinement, helps suppress performance degradation when the provided history is unreliable.\color{black}

\subsection{Data Selection and Inference Scaling for DPO}

We conducted an ablation study on the WER selection threshold (based on the $N=2$, 0.5 dropout SFT model). As shown in Table \ref{tab:DPO alpha and Data Selection} in Appendix \ref{sec:dif bad case}, our method is remarkably robust to data strictness, yielding consistent gains across all thresholds without requiring precise tuning. Furthermore, the optimal inference scaling $\gamma$ remains stable in all of these configurations, consistent with Sec. \ref{sec: Impact of DPO Inference Scaling}. This shows that the optimal inference strategy is effectively decoupled from the data curation process, significantly simplifying deployment.


\section{Conclusion}
In this work, we address contextual exposure bias in Speech-LLMs by proposing a unified training framework that integrates Teacher Error Knowledge, context dropout, and DPO. We demonstrate that training with Teacher Error Knowledge effectively bridges the train-test gap, significantly outperforming oracle-trained baselines. Crucially, context dropout proves to be a decisive factor for stability; it prevents the model from over-relying on textual history. Furthermore, DPO explicitly suppresses error propagation, yielding superior in-domain performance and robust cross-domain generalization. Collectively, our approach establishes a reliable paradigm for long-form ASR under realistic, imperfect conditions.

\section*{Limitations}
While our proposed framework effectively mitigates contextual exposure bias and enhances robustness in Speech-LLMs, several limitations remain to be addressed in future work.

First, Multi-Speaker Overlap. Our current experimental setup assumes sequential, turn-based speech (as found in TED-LIUM 3 and LibriSpeech). We have not evaluated the model's performance in scenarios involving overlapping speech or "cocktail party" environments. Since Speech-LLMs process audio as a single sequence, handling simultaneous speakers with heavy overlap likely requires specialized architectural modifications or data curation strategies that are beyond the scope of this work.

Second, Limited Diversity of Teacher Error Sources. Although Teacher Error Knowledge is a model-agnostic concept that can, in principle, be instantiated with hypotheses from arbitrary ASR systems, our current implementation relies on a single teacher model (Whisper large-v3) to generate erroneous contextual knowledge. Consequently, the simulated errors exposed during training are biased toward the error characteristics of this teacher, and may not exhaustively represent the diverse failure modes of Speech-LLMs or the wide range of acoustic distortions encountered in unconstrained environments.
\bibliography{custom}

@inproceedings{ma2025speech,
  title={Speech Recognition Meets Large Language Model: Benchmarking, Models, and Exploration},
  author={Ma, Ziyang and Yang, Guanrou and Yang, Yifan and Gao, Zhifu and Wang, Jiaming and Du, Zhihao and Yu, Fan and Chen, Qian and Zheng, Siqi and Zhang, Shiliang and others},
  booktitle={Proceedings of the AAAI Conference on Artificial Intelligence},
  volume={39},
  pages={24840--24848},
  year={2025}
}

@article{li2022recent,
  title={Recent advances in end-to-end automatic speech recognition},
  author={Li, Jinyu and others},
  journal={APSIPA Transactions on Signal and Information Processing},
  volume={11},
  year={2022},
  publisher={Now Publishers, Inc.}
}

@inproceedings{graves2006connectionist,
  title={Connectionist temporal classification: labelling unsegmented sequence data with recurrent neural networks},
  author={Graves, Alex and Fern{\'a}ndez, Santiago and Gomez, Faustino and Schmidhuber, J{\"u}rgen},
  booktitle={Proceedings of the 23rd international conference on Machine learning},
  pages={369--376},
  year={2006}
}

@inproceedings{chan2016listen,
  title={Listen, attend and spell: A neural network for large vocabulary conversational speech recognition},
  author={Chan, William and Jaitly, Navdeep and Le, Quoc and Vinyals, Oriol},
  booktitle={2016 IEEE international conference on acoustics, speech and signal processing (ICASSP)},
  pages={4960--4964},
  year={2016},
  organization={IEEE}
}

@inproceedings{graves2013speech,
  title={Speech recognition with deep recurrent neural networks},
  author={Graves, Alex and Mohamed, Abdel-rahman and Hinton, Geoffrey},
  booktitle={2013 IEEE international conference on acoustics, speech and signal processing},
  pages={6645--6649},
  year={2013},
  organization={Ieee}
}

@article{hsu2021hubert,
  title={Hubert: Self-supervised speech representation learning by masked prediction of hidden units},
  author={Hsu, Wei-Ning and Bolte, Benjamin and Tsai, Yao-Hung Hubert and Lakhotia, Kushal and Salakhutdinov, Ruslan and Mohamed, Abdelrahman},
  journal={IEEE/ACM transactions on audio, speech, and language processing},
  volume={29},
  pages={3451--3460},
  year={2021},
  publisher={IEEE}
}

@inproceedings{radford2023robust,
  title={Robust speech recognition via large-scale weak supervision},
  author={Radford, Alec and Kim, Jong Wook and Xu, Tao and Brockman, Greg and McLeavey, Christine and Sutskever, Ilya},
  booktitle={International conference on machine learning},
  pages={28492--28518},
  year={2023},
  organization={PMLR}
}

@article{cui2024recent,
  title={Recent advances in speech language models: A survey},
  author={Cui, Wenqian and Yu, Dianzhi and Jiao, Xiaoqi and Meng, Ziqiao and Zhang, Guangyan and Wang, Qichao and Guo, Yiwen and King, Irwin},
  journal={arXiv preprint arXiv:2410.03751},
  year={2024}
}

@inproceedings{aleksic2015bringing,
  title={Bringing contextual information to google speech recognition},
  author={Aleksic, Petar and Ghodsi, Mohammadreza and Michaely, Assaf and Allauzen, Cyril and Hall, Keith and Roark, Brian and Rybach, David and Moreno, Pedro},
  booktitle={Proc. Interspeech 2015},
  pages={468--472},
  year={2015}
}

@inproceedings{hall2015composition,
  title={Composition-based on-the-fly rescoring for salient n-gram biasing},
  author={Hall, Keith and Cho, Eunjoon and Allauzen, Cyril and Beaufays, Fran{\c{c}}oise and Coccaro, Noah and Nakajima, Kaisuke and Riley, Michael and Roark, Brian and Rybach, David and Zhang, Linda},
  booktitle={Proc. Interspeech 2015},
  pages={1418--1422},
  year={2015}
}

@article{fang2025joint,
  title={Joint decoding method for controllable contextual speech recognition based on Speech LLM},
  author={Fang, Yangui and Peng, Jing and Xi, Yu and Li, Xu and Li, Haoyu and Zhang, Chengwei and Zhong, Guohui and Yu, Kai},
  journal={arXiv preprint arXiv:2508.08585},
  year={2025}
}

@inproceedings{mcdermott2019density,
  title={A density ratio approach to language model fusion in end-to-end automatic speech recognition},
  author={McDermott, Erik and Sak, Hasim and Variani, Ehsan},
  booktitle={2019 IEEE Automatic Speech Recognition and Understanding Workshop (ASRU)},
  pages={434--441},
  year={2019},
  organization={IEEE}
}

@article{ravi2020improving,
  title={Improving accuracy of rare words for rnn-transducer through unigram shallow fusion},
  author={Ravi, Vijay and Gu, Yile and Gandhe, Ankur and Rastrow, Ariya and Liu, Linda and Filimonov, Denis and Novotney, Scott and Bulyko, Ivan},
  journal={arXiv preprint arXiv:2012.00133},
  year={2020}
}

@inproceedings{toshniwal2018comparison,
  title={A comparison of techniques for language model integration in encoder-decoder speech recognition},
  author={Toshniwal, Shubham and Kannan, Anjuli and Chiu, Chung-Cheng and Wu, Yonghui and Sainath, Tara N and Livescu, Karen},
  booktitle={2018 IEEE spoken language technology workshop (SLT)},
  pages={369--375},
  year={2018},
  organization={IEEE}
}

@article{wang2024deep,
  title={Deep CLAS: Deep Contextual Listen, Attend and Spell},
  author={Wang, Mengzhi and Xiong, Shifu and Wan, Genshun and Chen, Hang and Gao, Jianqing and Dai, Lirong},
  journal={arXiv preprint arXiv:2409.17603},
  year={2024}
}

@inproceedings{lakomkin2024end,
  title={End-to-end speech recognition contextualization with large language models},
  author={Lakomkin, Egor and Wu, Chunyang and Fathullah, Yassir and Kalinli, Ozlem and Seltzer, Michael L and Fuegen, Christian},
  booktitle={ICASSP 2024-2024 IEEE International Conference on Acoustics, Speech and Signal Processing (ICASSP)},
  pages={12406--12410},
  year={2024},
  organization={IEEE}
}

@article{li2024rag,
  title={LA-RAG: Enhancing LLM-based ASR Accuracy with Retrieval-Augmented Generation},
  author={Li, Shaojun and Shang, Hengchao and Wei, Daimeng and Guo, Jiaxin and Li, Zongyao and He, Xianghui and Zhang, Min and Yang, Hao},
  journal={CoRR},
  year={2024}
}

@inproceedings{chen2024salm,
  title={Salm: Speech-augmented language model with in-context learning for speech recognition and translation},
  author={Chen, Zhehuai and Huang, He and Andrusenko, Andrei and Hrinchuk, Oleksii and Puvvada, Krishna C and Li, Jason and Ghosh, Subhankar and Balam, Jagadeesh and Ginsburg, Boris},
  booktitle={ICASSP 2024-2024 IEEE International Conference on Acoustics, Speech and Signal Processing (ICASSP)},
  pages={13521--13525},
  year={2024},
  organization={IEEE}
}

@article{yang2024mala,
  title={MaLa-ASR: Multimedia-Assisted LLM-Based ASR},
  author={Yang, Guanrou and Ma, Ziyang and Yu, Fan and Gao, Zhifu and Zhang, Shiliang and Chen, Xie},
  journal={CoRR},
  year={2024}
}

@article{shen2025retrieval,
  title={Retrieval-Augmented Speech Recognition Approach for Domain Challenges},
  author={Shen, Peng and Lu, Xugang and Kawai, Hisashi},
  journal={arXiv preprint arXiv:2502.15264},
  year={2025}
}

@article{huang2025step,
  title={Step-Audio: Unified Understanding and Generation in Intelligent Speech Interaction},
  author={Huang, Ailin and Wu, Boyong and Wang, Bruce and Yan, Chao and Hu, Chen and Feng, Chengli and Tian, Fei and Shen, Feiyu and Li, Jingbei and Chen, Mingrui and others},
  journal={CoRR},
  year={2025}
}

@inproceedings{hernandez2018ted,
  author="Hernandez, Fran{\c{c}}ois
  and Nguyen, Vincent
  and Ghannay, Sahar
  and Tomashenko, Natalia
  and Est{\`e}ve, Yannick",
  title="TED-LIUM 3: Twice as Much Data and Corpus Repartition for Experiments on Speaker Adaptation",
  booktitle="Speech and Computer",
  year="2018",
  publisher="Springer International Publishing",
  pages="198--208",
}

@inproceedings{hulora,
  title={LoRA: Low-Rank Adaptation of Large Language Models},
  author={Hu, Edward J and Wallis, Phillip and Allen-Zhu, Zeyuan and Li, Yuanzhi and Wang, Shean and Wang, Lu and Chen, Weizhu and others},
  booktitle={International Conference on Learning Representations},
  year={2021}
}

@inproceedings{guo2023improved,
  title={Improved keyword recognition based on aho-corasick automaton},
  author={Guo, Yachao and Qiu, Zhibin and Huang, Hao and Siong, Chng Eng},
  booktitle={2023 International Joint Conference on Neural Networks (IJCNN)},
  pages={1--7},
  year={2023},
  organization={IEEE}
}

@article{zheng2023judging,
  title={Judging llm-as-a-judge with mt-bench and chatbot arena},
  author={Zheng, Lianmin and Chiang, Wei-Lin and Sheng, Ying and Zhuang, Siyuan and Wu, Zhanghao and Zhuang, Yonghao and Lin, Zi and Li, Zhuohan and Li, Dacheng and Xing, Eric and others},
  journal={Advances in neural information processing systems},
  volume={36},
  pages={46595--46623},
  year={2023}
}

@inproceedings{panayotov2015librispeech,
  title={Librispeech: an asr corpus based on public domain audio books},
  author={Panayotov, Vassil and Chen, Guoguo and Povey, Daniel and Khudanpur, Sanjeev},
  booktitle={2015 IEEE international conference on acoustics, speech and signal processing (ICASSP)},
  pages={5206--5210},
  year={2015},
  organization={IEEE}
}

@inproceedings{gao2023scaling,
  title={Scaling laws for reward model overoptimization},
  author={Gao, Leo and Schulman, John and Hilton, Jacob},
  booktitle={International Conference on Machine Learning},
  pages={10835--10866},
  year={2023},
  organization={PMLR}
}

@article{jiang2025huvidpo,
  title={Huvidpo: Enhancing video generation through direct preference optimization for human-centric alignment},
  author={Jiang, Lifan and Wu, Boxi and Zhang, Jiahui and Guan, Xiaotong and Chen, Shuang},
  journal={arXiv preprint arXiv:2502.01690},
  year={2025}
}

@article{cheng2024context,
  title={Context-aware speech recognition using prompts for language learners},
  author={Cheng, Jian},
  journal={Interspeech 2024},
  pages={4009--4013},
  year={2024}
}

@inproceedings{tang2024improving,
  title={Improving ASR contextual biasing with guided attention},
  author={Tang, Jiyang and Kim, Kwangyoun and Shon, Suwon and Wu, Felix and Sridhar, Prashant},
  booktitle={ICASSP 2024-2024 IEEE International Conference on Acoustics, Speech and Signal Processing (ICASSP)},
  pages={12096--12100},
  year={2024},
  organization={IEEE}
}

@inproceedings{lei2025contextualization,
  title={Contextualization of ASR with LLM using phonetic retrieval-based augmentation},
  author={Lei, Zhihong and Na, Xingyu and Xu, Mingbin and Pusateri, Ernest and Van Gysel, Christophe and Zhang, Yuanyuan and Han, Shiyi and Huang, Zhen},
  booktitle={ICASSP 2025-2025 IEEE International Conference on Acoustics, Speech and Signal Processing (ICASSP)},
  pages={1--5},
  year={2025},
  organization={IEEE}
}

@article{huang2024optimizing,
  title={Optimizing large-scale context retrieval for end-to-end ASR},
  author={Huang, Zhiqi and Caseiro, Diamantino and Joshi, Kandarp and Li, Christopher and Rondon, Pat and Wu, Zelin and Zadrazil, Petr and Zhou, Lillian},
  journal={Proc. Interspeech. ISCA},
  pages={4573--4577},
  year={2024}
}

@inproceedings{huang2024improving,
  title={Improving Neural Biasing for Contextual Speech Recognition by Early Context Injection and Text Perturbation},
  author={Huang, Ruizhe and Yarmohammadi, Mahsa and Khudanpur, Sanjeev and Povey, Daniel},
  booktitle={Proc. Interspeech 2024},
  pages={752--756},
  year={2024}
}

@article{kolokolov2024self,
  title={Self-consistent context aware conformer transducer for speech recognition},
  author={Kolokolov, Konstantin and Pekichev, Pavel and Raghunathan, Karthik},
  journal={arXiv preprint arXiv:2402.06592},
  year={2024}
}

@inproceedings{shi2024seaco,
  title={SeACo-Paraformer: A non-autoregressive ASR system with flexible and effective hotword customization ability},
  author={Shi, Xian and Yang, Yexin and Li, Zerui and Chen, Yanni and Gao, Zhifu and Zhang, Shiliang},
  booktitle={ICASSP 2024-2024 IEEE International Conference on Acoustics, Speech and Signal Processing (ICASSP)},
  pages={10346--10350},
  year={2024},
  organization={IEEE}
}

@inproceedings{koshkin2024llms,
  title={LLMs Are Zero-Shot Context-Aware Simultaneous Translators},
  author={Koshkin, Roman and Sudoh, Katsuhito and Nakamura, Satoshi},
  booktitle={Proceedings of the 2024 Conference on Empirical Methods in Natural Language Processing},
  pages={1192--1207},
  year={2024}
}

@inproceedings{sudo2024contextualized,
  title={Contextualized automatic speech recognition with dynamic vocabulary},
  author={Sudo, Yui and Fukumoto, Yosuke and Shakeel, Muhammad and Peng, Yifan and Watanabe, Shinji},
  booktitle={2024 IEEE Spoken Language Technology Workshop (SLT)},
  pages={78--85},
  year={2024},
  organization={IEEE}
}

@article{gong2024contextual,
  title={Contextual biasing speech recognition in speech-enhanced large language model},
  author={Gong, Xun and Lv, Anqi and Wang, Zhiming and Qian, Yanmin},
  journal={Proc. Interspeech. ISCA},
  pages={257--261},
  year={2024}
}

@article{zhou2025boosting,
  title={Boosting Context-Aware Speech Translation With Large Language Models},
  author={Zhou, Yue and Yuan, Yuxuan and Zhang, Chengwei and Shi, Xiaodong},
  journal={IEEE Signal Processing Letters},
  year={2025},
  publisher={IEEE}
}

@article{mu2025hearing,
  title={Hearing More with Less: Multi-Modal Retrieval-and-Selection Augmented Conversational LLM-Based ASR},
  author={Mu, Bingshen and Liu, Hexin and Xue, Hongfei and Wei, Kun and Xie, Lei},
  journal={arXiv preprint arXiv:2508.01166},
  year={2025}
}

@inproceedings{gourav2024multi,
  title={Multi-modal retrieval for large language model based speech recognition},
  author={Gourav, Aditya and Kolehmainen, Jari and Shivakumar, Prashanth and Gu, Yile and Strimel, Grant and Gandhe, Ankur and Rastrow, Ariya and Bulyko, Ivan},
  booktitle={Findings of the Association for Computational Linguistics ACL 2024},
  pages={4435--4446},
  year={2024}
}

@article{yusuyin2025whistle,
  title={Whistle: Data-efficient multilingual and crosslingual speech recognition via weakly phonetic supervision},
  author={Yusuyin, Saierdaer and Ma, Te and Huang, Hao and Zhao, Wenbo and Ou, Zhijian},
  journal={IEEE Transactions on Audio, Speech and Language Processing},
  year={2025},
  publisher={IEEE}
}

@article{ouyang2022training,
  title={Training language models to follow instructions with human feedback},
  author={Ouyang, Long and Wu, Jeffrey and Jiang, Xu and Almeida, Diogo and Wainwright, Carroll and Mishkin, Pamela and Zhang, Chong and Agarwal, Sandhini and Slama, Katarina and Ray, Alex and others},
  journal={Advances in neural information processing systems},
  volume={35},
  pages={27730--27744},
  year={2022}
}

@article{rafailov2023direct,
  title={Direct preference optimization: Your language model is secretly a reward model},
  author={Rafailov, Rafael and Sharma, Archit and Mitchell, Eric and Manning, Christopher D and Ermon, Stefano and Finn, Chelsea},
  journal={Advances in neural information processing systems},
  volume={36},
  pages={53728--53741},
  year={2023}
}

@inproceedings{jain2020contextual,
  title={Contextual RNN-T for Open Domain ASR},
  author={Jain, Mahaveer and Keren, Gil and Mahadeokar, Jay and Zweig, Geoffrey and Metze, Florian and Saraf, Yatharth},
  booktitle={Proc. Interspeech 2020},
  pages={11--15},
  year={2020}
}

@inproceedings{hou2022bring,
  title={Bring dialogue-context into RNN-T for streaming ASR.},
  author={Hou, Junfeng and Chen, Jinkun and Li, Wanyu and Tang, Yufeng and Zhang, Jun and Ma, Zejun},
  booktitle={INTERSPEECH},
  pages={2048--2052},
  year={2022}
}

@inproceedings{hori2020transformer,
  title={Transformer-Based Long-Context End-to-End Speech Recognition.},
  author={Hori, Takaaki and Moritz, Niko and Hori, Chiori and Le Roux, Jonathan},
  booktitle={Interspeech},
  pages={5011--5015},
  year={2020}
}

@inproceedings{hussein2024enhancing,
  title={Enhancing end-to-end conversational speech translation through target language context utilization},
  author={Hussein, Amir and Yan, Brian and Anastasopoulos, Antonios and Watanabe, Shinji and Khudanpur, Sanjeev},
  booktitle={ICASSP 2024-2024 IEEE International Conference on Acoustics, Speech and Signal Processing (ICASSP)},
  pages={11971--11975},
  year={2024},
  organization={IEEE}
}

@inproceedings{lee2024enhancing,
  title={Enhancing Dialogue Speech Recognition with Robust Contextual Awareness via Noise Representation Learning},
  author={Lee, Wonjun and Kim, San and Lee, Gary Geunbae},
  booktitle={Proceedings of the 25th Annual Meeting of the Special Interest Group on Discourse and Dialogue},
  pages={333--343},
  year={2024}
}

\appendix

\section{Data Selection and Inference Scaling for DPO}
\label{sec:dif bad case}

\begin{table*}[t]
    \centering
    \setlength{\tabcolsep}{5.5pt}
    \begin{tabular}{c|l|ccc|ccc}
        \hline
         \multirow{2}{*}{\textbf{WER (\%) Threshold}} & \multirow{2}{*}{\textbf{$\gamma$}} & \multicolumn{3}{c|}{\textbf{TED-LIUM 3 (WER \%) $\downarrow$}} & \multicolumn{3}{c}{\textbf{LibriSpeech (WER \%) $\downarrow$}} \\
         & & \textbf{Attacks/o} & \textbf{Attacks/w} & \textbf{Gap $\downarrow$} & \textbf{Test-clean} & \textbf{Test-other} & \textbf{Ave.} \\
        \hline
        \multirow{8}{*}{5} & 0 & 5.47 & 7.93 & 2.46 & 5.14 & 9.50 & 7.320 \\
        & 0.0625  & 5.40 & 7.02 & 1.62 & 5.11 & \textbf{9.32} & \textbf{7.215} \\
        & 0.125  & 5.43 & 7.34 & 1.91 & 5.18 & 9.41 & 7.295 \\
        & 0.1875  & \textbf{5.04} & 6.76 & 1.72 & \textbf{5.09} & 9.70 & 7.395 \\
        & 0.25  & 5.23 & \textbf{5.99} & 0.76 & 5.18 & 9.51 & 7.345 \\
        & 0.375 & 6.21 & 6.42 & \textbf{0.21} & 5.49 & 10.19 & 7.840 \\
        & 0.5 & 9.18 & 9.63 & 0.45 & 8.58 & 14.65 & 11.615 \\
        & 0.625 & 50.40 & 50.97 & 0.57 & 69.98 & 70.66 & 70.320 \\
        \hline

        \multirow{8}{*}{10} & 0 & 5.47 & 7.93 & 2.46 & 5.14 & 9.50 & 7.320 \\
        & 0.0625  & 5.40  & 7.10  & 1.70 & 5.11  & \textbf{9.30}  & 7.205  \\
        & 0.125  & 5.32  & 6.91  & 1.59 & 5.19  & 9.61  & 7.400  \\
        & 0.1875  & \textbf{5.04}  & 6.63  & 1.59 & 5.04  & 9.67  & 7.355  \\
        & 0.25  & 5.27  & \textbf{6.00}  & 0.73 & \textbf{4.62}  & 9.50  & \textbf{7.060}  \\
        & 0.375 & 6.29  & 6.38  & \textbf{0.09} & 5.06  & 10.22 & 7.640  \\
        & 0.5 & 10.22 & 10.87 & 0.65 & 8.65  & 14.96 & 11.805 \\
        & 0.625 & 85.55 & 89.97 & 4.42 & 88.93 & 85.49 & 87.210 \\
        \hline

        \multirow{8}{*}{15} & 0 & 5.47 & 7.93 & 2.46 & 5.14 & 9.50 & 7.320 \\
        & 0.0625  & 5.40  & 7.43  & 2.03 & 5.12  & 9.30  & 7.210  \\
        & 0.125  & 5.15  & 6.63  & 1.48 & 5.21  & 9.58  & 7.395  \\
        & 0.1875  & 5.12  & 5.91  & 0.79 & 4.98  & 9.46  & 7.220  \\
        & 0.25  & \textbf{5.07}  & \textbf{5.67}  & \textbf{0.60} & \textbf{4.77}  & \textbf{9.27}  & \textbf{7.020}  \\
        & 0.375 & 5.79  & 6.14  & 0.35 & 5.04  & 9.84  & 7.440  \\
        & 0.5 & 8.27  & 8.84  & 0.57 & 6.99  & 12.89 & 9.940  \\
        & 0.625 & 33.58 & 35.59 & 2.01 & 22.72 & 27.96 & 25.340 \\
        \hline

        \multirow{8}{*}{20} & 0  & 5.47  & 7.93  & 2.46 & 5.14 & 9.50  & 7.320  \\
        & 0.0625  & 5.37  & 7.13  & 1.76 & 5.12 & 9.31  & 7.215  \\
        & 0.125  & 5.11  & 5.76  & 0.65 & 5.02 & 9.53  & 7.275  \\
        & 0.1875  & \textbf{5.06} & 5.69 & 0.63 & \textbf{4.70} & \textbf{9.08} & \textbf{6.890} \\
        & 0.25  & 5.17  & \textbf{5.63} & 0.46 & 4.84 & 9.19  & 7.015  \\
        & 0.375 & 5.55  & 5.73  & \textbf{0.18} & 4.85 & 9.63  & 7.240  \\
        & 0.5 & 8.39  & 8.67  & 0.28 & 6.44 & 12.14 & 9.290  \\
        & 0.625 & 53.26 & 57.15 & 3.89 & 27.11 & 28.96 & 28.035 \\
        \hline

        \multirow{8}{*}{25} & 0  & 5.47  & 7.93  & 2.46 & 5.14 & 9.50  & 7.320  \\
        & 0.0625  & 5.39  & 7.40  & 2.01  & 5.13  & 9.33  & 7.230  \\
        & 0.125  & 5.28  & 6.00  & 0.72  & 5.21  & 9.57  & 7.390  \\
        & 0.1875  & \textbf{5.08}  & 5.68  & 0.60  & \textbf{4.70}  & \textbf{9.18}  & \textbf{6.940}  \\
        & 0.25  & 5.28  & \textbf{5.34}  & \textbf{0.06}  & 4.84  & 9.33  & 7.085  \\
        & 0.375 & 5.74  & 6.04  & 0.30  & 5.12  & 10.01 & 7.565  \\
        & 0.5 & 10.26 & 10.50 & 0.24  & 8.25  & 14.64 & 11.445 \\
        & 0.625 & 31.01 & 30.48 & -0.53 & 28.03 & 35.84 & 31.935 \\
        \hline

        \multirow{8}{*}{30} & 0  & 5.47  & 7.93  & 2.46 & 5.14 & 9.50  & 7.320  \\
        & 0.0625  & 5.32 & 7.38 & 2.06 & 5.14 & \textbf{9.28}  & 7.210  \\
        & 0.125  & 5.12 & 6.06 & 0.94 & 5.14 & 9.48  & 7.310  \\
        & 0.1875  & 5.00 & 5.51 & 0.51 & \textbf{4.70} & 9.54  & 7.120  \\
        & 0.25  & \textbf{4.97} & 5.39 & 0.42 & 4.74 & \textbf{9.28}  & \textbf{7.010}  \\
        & 0.375 & \textbf{4.97} & \textbf{5.20} & \textbf{0.23} & 4.83 & 9.33  & 7.080  \\
        & 0.5 & 5.49 & 5.78 & 0.29 & 5.11 & 10.00 & 7.555  \\
        & 0.625 & 8.77 & 9.11 & 0.34 & 8.16 & 14.73 & 11.445 \\
        \hline
    \end{tabular}
    \caption{Impact of Hard Negatives threshold and DPO LoRA scaling factor ($\gamma$) during inference. TED-LIUM 3 Gap denotes the WER degradation caused by irrelevant context attacks.}
    \label{tab:DPO alpha and Data Selection}
\end{table*}

\end{document}